\documentclass[letterpaper, 10 pt, conference]{ieeeconf}  % Comment this line out if you need a4paper

\IEEEoverridecommandlockouts                              % This command is only needed if 
\usepackage{graphics} 
\usepackage{epsfig} 
\usepackage{mathptmx} 
\usepackage{times} 
\usepackage{amsmath} 
\usepackage{amssymb}  
\usepackage[T1]{fontenc}
\usepackage{stfloats}
\usepackage{threeparttable}
\usepackage{soul, color,xcolor}
\usepackage{booktabs}
\usepackage{url}
\usepackage{cite}

\title{\LARGE \bf
ReJSHand: Efficient Real-Time Hand Pose Estimation and Mesh Reconstruction Using Refined Joint and Skeleton Features
}

\author{Shan An$^{1}$, \textit{Senior Member, IEEE}, Shipeng Dai$^{2}$, Mahrukh Ansari$^{1}$,  Yu Liang$^{3}$, Ming Zeng$^{1,*}$, \\  Konstantinos A. Tsintotas$^{4}$, \textit{Senior Member, IEEE},
Changhong Fu$^{5}$, 
Hong Zhang$^{6}$, \textit{Fellow, IEEE}
\thanks{$^{1}$Shan An, Mahrukh Ansari, and Ming Zeng are with the School of Electrical and Information Engineering, Tianjin University, Tianjin 300072, China {\tt\small $\lbrace$anshan, Mahrukh3,zengming$\rbrace$@tju.edu.cn}}%
\thanks{$^{2}$Shipeng Dai is with the College of Sciences, Northeastern University, Shenyang 110004, China {\tt\small neu\_daishipeng@163.com}}
\thanks{$^{3}$Yu Liang is with the College of Computer Science, Beijing University of Technology, Beijing 100124, China {\tt\small yuliang@bjut.edu.cn}}
\thanks{$^{4}$Konstantinos A. Tsintotas is with the Department of Production and Management Engineering, Democritus
University of Thrace, Xanthi 67132, Greece {\tt\footnotesize ktsintot@pme.duth.gr} }%
\thanks{$^{5}$Changhong Fu is with the School of Mechanical Engineering, Tongji University, Shanghai 201804, China {\tt\small changhongfu@tongji.edu.cn}       }%	  
\thanks{$^{6}$Hong Zhang is with the Department of Electronic and Electrical Engineering, Southern University of Science and Technology, Shenzhen 518055, China {\tt\small hzhang@ualberta.ca}        }%
\thanks{$^{*}$Corresponding Author}
}
\begin{document}

\maketitle
\thispagestyle{empty}
\pagestyle{empty}

%%%%%%%%%%%%%%%%%%%%%%%%%%%%%%%%%%%%%%%%%%%%%%%%%%%%%%%%%%%%%%%%%%%%%%%%%%%%%%%%

\begin{abstract}

Accurate hand pose estimation is vital in robotics, advancing dexterous manipulation in human-computer interaction. Toward this goal, this paper presents ReJSHand (which stands for Refined Joint and Skeleton Features), a cutting-edge network formulated for real-time hand pose estimation and mesh reconstruction.
The proposed framework is designed to accurately predict 3D hand gestures under real-time constraints, which is essential for systems that demand agile and responsive hand motion tracking. The network's design prioritizes computational efficiency without compromising accuracy, a prerequisite for instantaneous robotic interactions. Specifically, ReJSHand comprises a 2D keypoint generator, a 3D keypoint generator, an expansion block, and a feature interaction block for meticulously reconstructing 3D hand poses from 2D imagery.
In addition, the multi-head self-attention mechanism and a coordinate attention layer enhance feature representation, streamlining the creation of hand mesh vertices through sophisticated feature mapping and linear transformation.
Regarding performance, comprehensive evaluations on the FreiHand dataset demonstrate ReJSHand's computational prowess. It achieves a frame rate of 72 frames per second while maintaining a PA-MPJPE (Position-Accurate Mean Per Joint Position Error) of 6.3 mm and a PA-MPVPE (Position-Accurate Mean Per Vertex Position Error) of 6.4 mm. Moreover, our model reaches scores of 0.756 for F@05 and 0.984 for F@15, surpassing modern pipelines and solidifying its position at the forefront of robotic hand pose estimators.
To facilitate future studies, we provide our source code at ~\url{https://github.com/daishipeng/ReJSHand}.

\end{abstract}

%%%%%%%%%%%%%%%%%%%%%%%%%%%%%%%%%%%%%%%%%%%%%%%%%%%%%%%%%%%%%

\section{Introduction}

\begin{figure}[t!]
    \centering
    \includegraphics[width=0.5\textwidth]{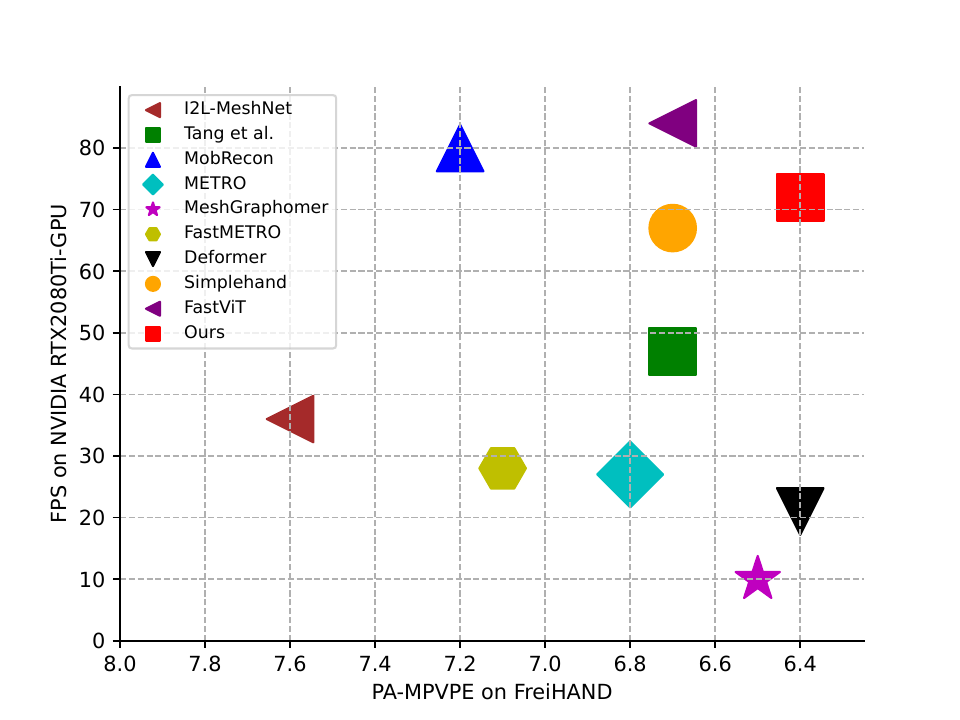}
    \caption{We compare hand pose estimators in terms of their accuracy and computational efficiency. ReJSHand achieves the best balance between accuracy and speed. All tests were conducted on an NVIDIA 2080Ti GPU.}
    \label{fig:FPS}
\end{figure}

The collection of demonstration data for robotic dexterous manipulation typically relies on expensive equipment. Vision-based methods can estimate 3D hand poses using only monocular cameras. By mapping the predicted hand joint positions to the trajectories of dexterous hand movements, vision-based approaches provide a feasible way to collect dexterous manipulation demonstration data at a low cost~\cite{qin2022one,fu2024humanplus}.
Moreover, with the widespread use of low-cost RGB consumer-grade cameras, hand pose and hand mesh recovery estimators are extensively studied for broad robotics applications~\cite{che2021detection,ohkawa2023efficient}.
%Augmented reality (AR) and virtual reality (VR) are among the applications that benefit the most~\cite{liebers2024identifying, buckingham2021hand}. In these applications, accurate hand pose estimation enables users to effectively control objects through gestures, thereby providing a means of interacting with digital content~\cite{che2021detection,ohkawa2023efficient}.

Nevertheless, accurately predicting 3D hand gestures while meeting real-time constraints remains an open challenge in robotics. To address this challenge, researchers have turned to lightweight network architectures that can operate efficiently with fewer computational resources. However, this approach presents a trade-off between complexity and efficiency, leading to the classification of hand pose estimators into two main categories.
The first category employs parametric models to predict hand shape and pose~\cite{romero2022embodied, xiao20233d}. While these methods are effective during prediction, they are highly complex due to their reliance on predefined parameter spaces derived from hand shape and pose datasets~\cite{liu2024deep}.
The second category consists of model-free, highly semantic strategies that directly predict vertex coordinates, capturing finer details of the hand. These techniques leverage Graph Convolutional Neural Networks (GCNNs) to simulate vertex-to-vertex interactions on triangular meshes~\cite{liu2024deep, chen2022mobrecon}. However, adding more parameters to enhance model performance leads to increased latency and larger model sizes~\cite{wang20243d}.

Nonetheless, despite existing pipelines in both categories demonstrating their ability to improve reconstruction quality, real-time performance remains a critical requirement. Toward this end, this paper proposes a method to map high-quality 3D hand models from images or video frames. Specifically, we introduce a lightweight keypoint detection technique that can respond quickly and accurately when tackling the task of hand pose estimation and mesh reconstruction.
The main contributions of our approach are as follows:

\begin{itemize}
    \item We introduce ReJSHand, a sophisticated network architecture for real-time hand pose estimation and mesh reconstruction using advanced computational methods.
    \item The expansion and feature interaction blocks are designed to meticulously refine joint and skeleton features. These blocks are essential for efficient hand pose recovery when input data originates from 2D imagery.
    \item We present a comprehensive experimental protocol to demonstrate its performance. ReJSHand achieves a frame rate of 72 frames per second, highlighting its suitability for dynamic and real-time interaction scenarios.
\end{itemize}

The rest of this paper is organized as follows. In Section~\ref{sec:related}, we review related work on hand pose estimation. Section~\ref{sec:methodology} details the proposed method, while the evaluation and experimental results are presented in Section~\ref{sec:experiments}. Finally, Section~\ref{sec:conclusions} concludes the paper.

\section{Related Work}
\label{sec:related}

%This section focuses on the most representative hand pose estimation and mesh reconstruction techniques when data is captured by a single camera. Hand pose estimation is essential as it enables real-time control and dexterous manipulations during teleoperation applications. Meanwhile, mesh reconstruction provides the detailed 3D mesh of the hand's predicted keypoints.
%A single-camera teleoperation system can achieve real-time control of customized robotic hands through hand pose estimation. 

\subsection{Hand pose estimation}

% It utilizes imitation learning to improve the performance and robustness of robots in carrying out complex manipulation tasks in the real world~\cite{qin2022one, qin2023anyteleop}.
To achieve dexterous manipulation, it is necessary to accurately capture the demonstration data of human hands, which requires precise hand pose estimation~\cite{qin2022one, qin2023anyteleop}. 2D keypoint prediction—identifying the 2D keypoints of the hand (such as hand joints, fingertips, and palm position)—is a crucial first step because it infers the 3D coordinates and solves the complete 3D hand pose, providing depth information and more spatial visualization along with camera parameters~\cite{chen2020nonparametric}.
Wei \textit{et al.}~\cite{wei2022dual} propose a dual regression approach and a lightweight network architecture for efficient hand pose estimation. Similarly, FastHand~\cite{an2022fasthand} adopts an encoder-decoder network and heatmap regression to achieve fast and accurate results on embedded devices. The real-time hand-tracking framework MediaPipe~\cite{zhang2020mediapipe} predicts a 2.5D hand pose from an input frame using a two-stage pipeline: a palm detector and a hand landmark model of the hand skeleton. However, this model struggles to capture intricate and occluded poses due to limitations in real-world and synthetic datasets~\cite{biswas2023mediapipe}.
In addition, Zimmermann \textit{et al.}~\cite{zimmermann2019freihand} introduce a dataset for markerless capture of hand pose and shape from a single image. This novel dataset includes 130,000 real-world images, facilitating research on hand pose estimation.

Another line of work focuses on lightweight CNNs that directly regress hand keypoint coordinates on a plane or a particular reference form for 2D and 3D hand pose prediction. This approach improves efficiency compared to traditional two-stage methods~\cite{santavas2020attention}. Similarly, self-attention mechanisms from transformers have been widely adopted in various fields, including hand pose estimation~\cite{vaswani2017attention}. For instance, METRO~\cite{lin2021end} is a multi-layer transformer encoder-based model that predicts 3D hand pose from a single image using masked vertex modeling, which enhances hand mesh reconstruction.
MeshGraphormers~\cite{lin2021mesh} introduce a linear multi-layer perceptron (MLP) technique and positional encoding that iteratively refines the coarse mesh to its original resolution, achieving state-of-the-art performance on multiple benchmarks. Another transformer-based architecture for 3D hand pose estimation employs a simple design, leveraging large-scale training datasets and model capacity~\cite{pavlakos2024reconstructing}. Although this framework demonstrates high performance, its major limitation lies in the substantial computational resources required, as it relies on extensive datasets that are not easily accessible. Additionally, its single-frame operation makes it slow for continuous scene tracking.
Lastly, Zhou \textit{et al.}~\cite{zhou2024simple} propose a simple method based on token generators and grid regressors for 3D hand pose estimation. However, due to their model settings, its performance requires further improvement.

\subsection{Hand mesh reconstruction}

3D hand reconstruction often relies on pre-trained model parameters to construct shape coefficients, such as SMPL~\cite{loper2023smpl} and MANO~\cite{zhou2020monocular}. However, numerous works have attempted to predict the hand mesh recovery coefficients directly~\cite{zhou2020monocular, moon2018v2v, moon2020i2l}. For example, Zhou \textit{et al.}~\cite{zhou2020monocular} estimate the MANO coefficients using a kinematic chain and an inverse kinematics model to reconstruct the pose and shape of the hand.
While SMPL and MANO can generate 3D meshes, they often incorporate 3D information into a volumetric space, where the 3D structure is less clear compared to explicit 3D vertices. Another strategy is based on Euclidean 3D representation, which directly applies canonical convolutional operators to voxels~\cite{moon2018v2v}. I2L-MeshNet methods further split the voxel into three-pixel spaces, producing meshes using a 2.5D approach~\cite{moon2020i2l}.
Despite utilizing Euclidean space to create hand meshes, these voxel-based and 2.5D methods are often inefficient and fail to capture the detailed 3D structure effectively.

In addition to parametric solutions, several authors have addressed the task of human body shape estimation by directly regressing the shape from input data~\cite{kolotouros2019convolutional, kushwaha20243dpmesh, li2024hhmr, jiang2024complementing}. For example, GraphCMR employs a Graph Convolutional Neural Network (GCNN) to regress 3D vertices directly from the input~\cite{kolotouros2019convolutional}.
Kushwaha \textit{et al.}~\cite{kushwaha20243dpmesh} propose a cascaded model for 3D mesh regression, which consists of three main components: 3D pose estimation, pose enhancement, and mesh articulation. These components work together to enhance overall mesh construction.
Li \textit{et al.}~\cite{li2024hhmr} introduce a graph diffusion model that improves multimodal fusion for holistic hand mesh reconstruction. Their method demonstrates the ability to reconstruct 3D hand meshes from noisy point cloud data.
Similarly, Jiang \textit{et al.}~\cite{jiang2024complementing} achieve high-quality 3D mesh construction using data from both event cameras and RGB cameras, which is particularly effective in complex outdoor settings.

\section{Methodology}
\label{sec:methodology}

\begin{figure*}[t!]
    \centering
    \includegraphics[width=\textwidth]{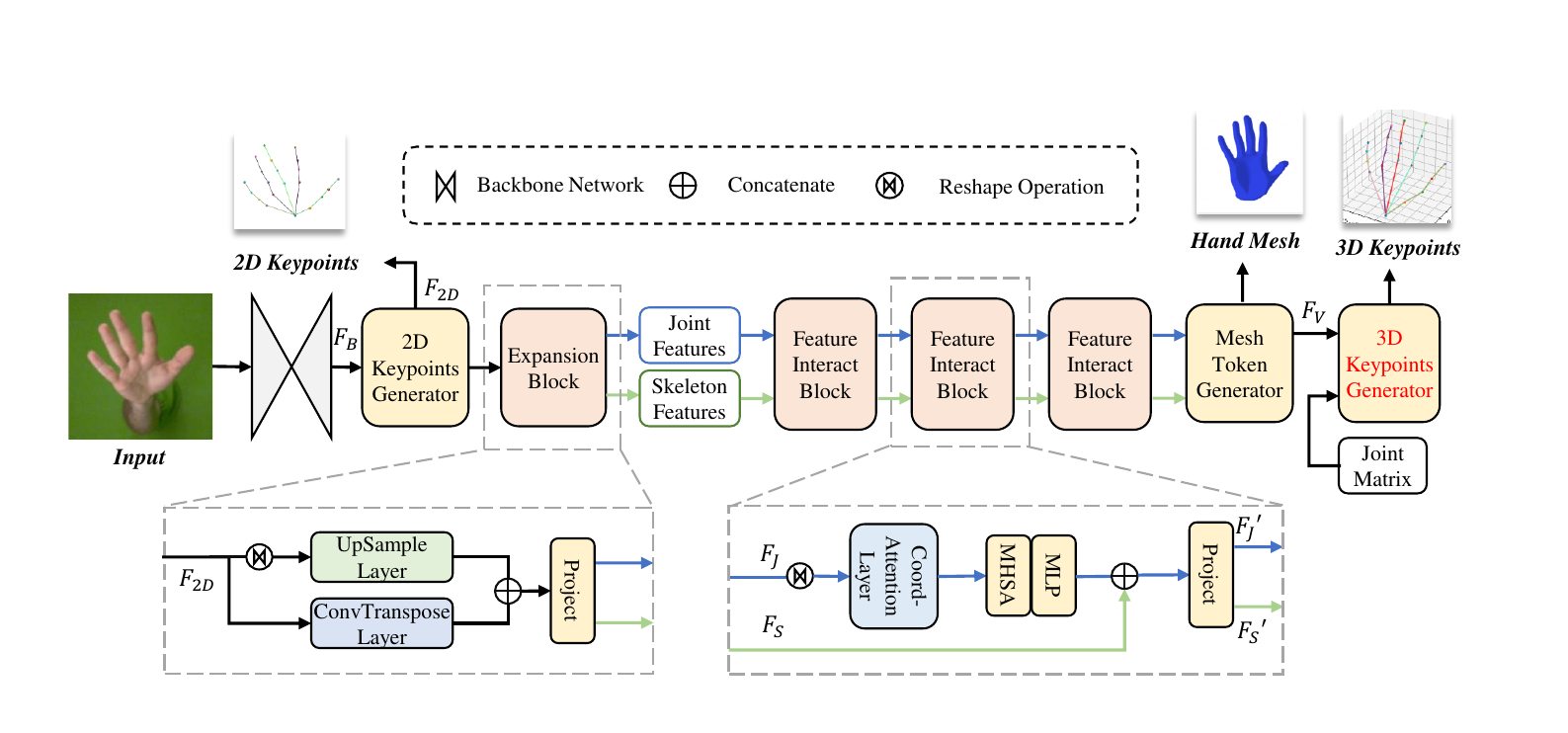}
    \caption{
    An overview of the proposed lightweight network for real-time hand pose estimation and mesh reconstruction, ReJSHand, is provided. First, the cropped hand images are processed through the backbone network to extract features. Next, the 2D keypoint generator maps these features to 2D coordinates. Simultaneously, the expansion block upsamples the feature map using transposed convolutional layers and sampling techniques. By jointly mapping both features, we leverage their complementary and synergistic roles in our hand pose estimator. The feature interaction block then refines the joint and skeleton features by learning coordinate dependencies through coordinate and multi-head attention modules. Subsequently, the mesh token generator integrates these refined features to generate mesh vertices. Finally, the 3D keypoint generator maps the mesh vertices to 2D keypoint coordinates by integrating the joint matrix.
    }
    \label{fig:ReJSHand}
\end{figure*}

\subsection{An overview of ReJSHand}

The network architecture of ReJSHand is depicted in Fig.~\ref{fig:ReJSHand}. 
The input to the proposed network is a cropped hand image $I\in~\mathbb{R}^{224\times224\times3}$, which is first processed by a pre-trained backbone network in order to extract the hand's features $F_B \in \mathbb{R} ^{640\times 7\times 7}$.
These features encapsulate the hand's primary characteristics and are directly fed into the 2D keypoints generator.
Next, through simple linear operations, this module produces the 2D keypoint positions of the hand in the image $F_{2D}\in\mathbb{R}^{21\times2}$.
Subsequently, $F_{2D}$ is placed into the expansion block, which generates joint features $F_J\in~\mathbb{R}^{21\times256}$ and skeleton features$F_S\in~\mathbb{R}^{21\times256}$.
The skeleton features, $F_S$, contain information about the connections between hand joints, which later guide the refined inference of joint features $F_J$.

$F_S$ and $F_J$ are then fed into the feature-interaction block, where the topological structure of the visible relationships between different hand joints is modeled through the interaction of $F_S$. After passing through three such blocks, the refined features $F_S'$ and $F_J'$ are fed into the mesh token generator. The mesh token $F_V \in \mathbb{R}^{778 \times 3}$ is obtained via concatenation and linear mapping. 
After rendering, $F_V$ generates a hand mesh that can be applied to tasks such as human-computer interaction. Subsequently, the 3D keypoints generator maps $F_V$ to 3D keypoints by fusing joint matrix information, ensuring accurate and robust keypoint detection. 
In summary, using a hand image as input, ReJSHand generates the hand's 2D keypoints, 3D keypoints, and mesh vertices. The following section introduces the detailed components mentioned earlier.

\subsection{Expansion Block}

Following the 2D hand pose estimation backbone from the previous phase, we designed a simple yet efficient expansion block to generate joint and skeleton features. Joint features refer to the position information of various hand joints, such as the wrist, finger joints, etc., in the image. These features aid the model in understanding the spatial relationships between different hand parts, which are particularly crucial for recovering 3D poses from monocular cameras. Skeleton features, on the other hand, refer to the connections between joints, providing clues about their relative positions, which help resolve depth ambiguities in monocular images.

The detailed process of the expansion block is as follows.
Initially, $F_B$ undergoes a transposed convolution operation, which is used for upsampling and resizing the feature maps.
This technique increases the input's spatial dimensions by applying a series of convolutional kernels, resulting in a larger feature map with higher resolution.
This operation aids in capturing image features at a more granular scale.
More importantly, the parameters required for transposed convolution are limited, making it suitable for lightweight networks.
Subsequently, the coordinate values in $F_{2D}$ are normalized to the range of [-1, 1], ensuring that these values can be accurately mapped onto the feature map's coordinate system, enabling precise sampling.
The adjusted $F_{2D}$ is then expanded to match the dimensions of the feature map after transposed convolution, preparing it for the sampling operation, as described in the following equation:
\begin{equation}
    F_{2D}' = (2F_{2D}-1)\otimes e,
\end{equation}
where $F_{2D}'$ is the upsampled feature, $\otimes$ denotes the expansion operation on the third dimension of $F_{2D}$, and $e$ is a vector with a dimension of 1.
Grid sampling techniques extract features from feature map based on the expanded coordinates of $F_{2D}$ .
This sampling process retrieves the corresponding feature values from the feature map according to the specified coordinates in $F_{2D}$.
Finally, $F_J$ and $F_S$ are obtained through linear mapping.

\subsection{Feature-Interaction Block}

The core of ReJSHand consists of three feature-interaction blocks, as they are connected in series to progressively refine $F_J$ and $F_S$, thereby generating accurate mesh tokens.
As shown in Fig.~\ref{fig:ReJSHand}, the joint features $F_J\in \mathbb{R}^{21\times256}$ (with the second and third block having joint feature dimensions of $84~\times128$ and $336~\times~64$, respectively) and the skeleton features $F_S~\in~\mathbb{R}^{21\times~256}$ (with the second and third blocks having skeleton feature dimensions of $84\times128$ and $336 \times~64$, respectively) are processed through the feature interact blocks.
First, $F_J$ is reshaped to match the expected input format of the coord-attention layer. This layer uses one-dimensional convolution operations to focus the network on important coordinates.
It processes $F_J$ while maintaining the feature dimensions and  achieves preliminary feature transformation through weight adjustment.
The dependencies within $F_J$ are fully captured using the multi-head self-attention mechanism~\cite{vaswani2017attention}.
Specifically, it first generates queries, keys, and values through three distinct linear layers (Eqn.~\ref{eq2}). Then, it calculates attention scores (Eqn.\ref{eq3}) based on these, which reflect the interrelationships between different coordinates in the sequence:
\begin{align}
    V_J = W_VF_J, K_J=W_KF_J, Q_J = W_QF_J,  \label{eq2} \\
    Att(Q_J, K_J ,V_J) = Softmax(\frac{Q_JK_J^T}{\sqrt{d_k}})V_J, \label{eq3} 
\end{align}
where $W_V, W_K$, and $W_Q$ are the weight matrices corresponding to the values, keys, and queries, respectively, and $d_k$ is the dimension of the key. 
The three feature-interaction blocks have $d_k$ values of 32, 16, and 8, respectively.
The number of heads is set to 8.
The output from the multi-head self-attention mechanism is further transformed through a linear layer to obtain the joint features. It is worth mentioning that these blocks enhance the expressiveness of the features and improve the learning capacity and stability of the network. 
Subsequently, this module outputs enriched and robust representations of $F_J'$ and $F_S'$ through a projection linear operation.

$F_J'$ and $F_S'$ play complementary and synergistic roles in 3D hand pose estimation. $F_J'$ provides precise localization information, while $F_S'$ captures the structural and topological relationships between hand joints. By combining these two types of features, the model can comprehensively understand the variations in hand pose, thereby achieving more accurate results across various application scenarios. The design of the feature interaction modules further enhances the model's performance, leading to high accuracy and robustness in hand pose estimation.

\subsection{Keypoints' Generator}

The hand image in Fig.~\ref{fig:ReJSHand} is first processed through a pre-trained backbone network, yielding a rich set of hand features denoted as $F_B$.
ReJSHand then directly performs a linear operation on $F_B$ to generate the coordinates of 2D keypoints, maintaining a high level of accuracy without a significant increase in computational cost.
For the 3D keypoints generator, the joint matrix must be multiplied and fused with $F_V~\in~\mathbb{R}^{778~\times~64}$.
This is a pre-trained parameter of the MANO model used to infer joint positions from vertex coordinates. Hand images captured by monocular cameras may contain a certain degree of occlusion, which is one of the leading causes of difficulty in hand pose estimation. The joint matrix learned from complete hand topological modeling helps reduce the impact of occlusions. It is worth noting that the officially published joint matrix only includes the mapping for 16 joint points, indicating that it does not cover the vertices of the five fingertips. Therefore, slicing these five points from the vertices is optional.
The choice to generate the hand mesh before generating 3D keypoints minimizes the loss. If the generation order is reversed, inaccurate 3D keypoints will result in an even more inaccurate hand mesh, thereby amplifying the loss.

After passing through the previous three feature-interaction blocks, $F_J'$ and $F_S'$ have already captured comprehensive hand joint and skeleton characteristics. 
Therefore, the mesh token generator can infer $F_V \in~\mathbb{R}^{778~\times~64}$ through concatenation and simple linear operations, thereby render a textured hand mesh.

\subsection{Loss Functions}

Based on the detailed descriptions above, ReJSHand generates the 2D keypoints' coordinates $J_{2D} \in~\mathbb{R}^{21~\times~2}$, the 3D keypoints' coordinates $J_{3D}~\in~\mathbb{R}^{21~\times 3}$, and the mesh vertices' coordinates $V~\in~\mathbb{R}^{778~\times~3}$ through supervised training. 
Given the true 2D keypoint coordinates $\hat{J}_{2D}$, the true 3D keypoint coordinates $\hat{J}_{3D}$, and the true mesh vertices' coordinates $\hat{V}$ from the image, the corresponding losses are defined as follows:
\begin{align}
    L_{J_{2D}} = \frac{\sum_{i=1}^{N_{2D}}| J_{2D}^{(i)} -\hat{J}_{2D}^{(i)} |}{N_{2D}}, \\
    L_{J_{3D}} = \frac{\sum_{i=1}^{N_{3D}}| J_{3D}^{(i)} -\hat{J}_{3D}^{(i)} |}{N_{3D}}, \\
    L_V = \frac{\sum_{i=1}^{N_{V}}| V^{(i)} -\hat{V}^{(i)} |}{N_V},
\end{align}
where $J_{2D}^{(i)}$ represents the \textit{i}-th pair of coordinates, with $N_{2D}$, $N_{3D}$, and $N_V$ being 21, 21, and 778, respectively. 

The final loss is defined as:
\begin{equation}
    L = k_{2D}L_{J_{2D}}+k_{3D}L_{J_{3D}} + k_VL_V.
\end{equation}
The weight coefficients $k_{2D}$, $k_{3D}$, and $k_V$ are used to balance the contributions of the different loss functions. 
These coefficients can be adjusted for various application tasks to determine the optimization direction.
Given that this paper aims to provide accurate hand meshes and 3D keypoint positions, the values of $k_{2D}$, $k_{3D}$, and $k_V$ are set to 1, 10, and 10, respectively.

\section{Experiments}
\label{sec:experiments}

\begin{figure*}[t!]
    \centering
    \includegraphics[width=0.91\textwidth]{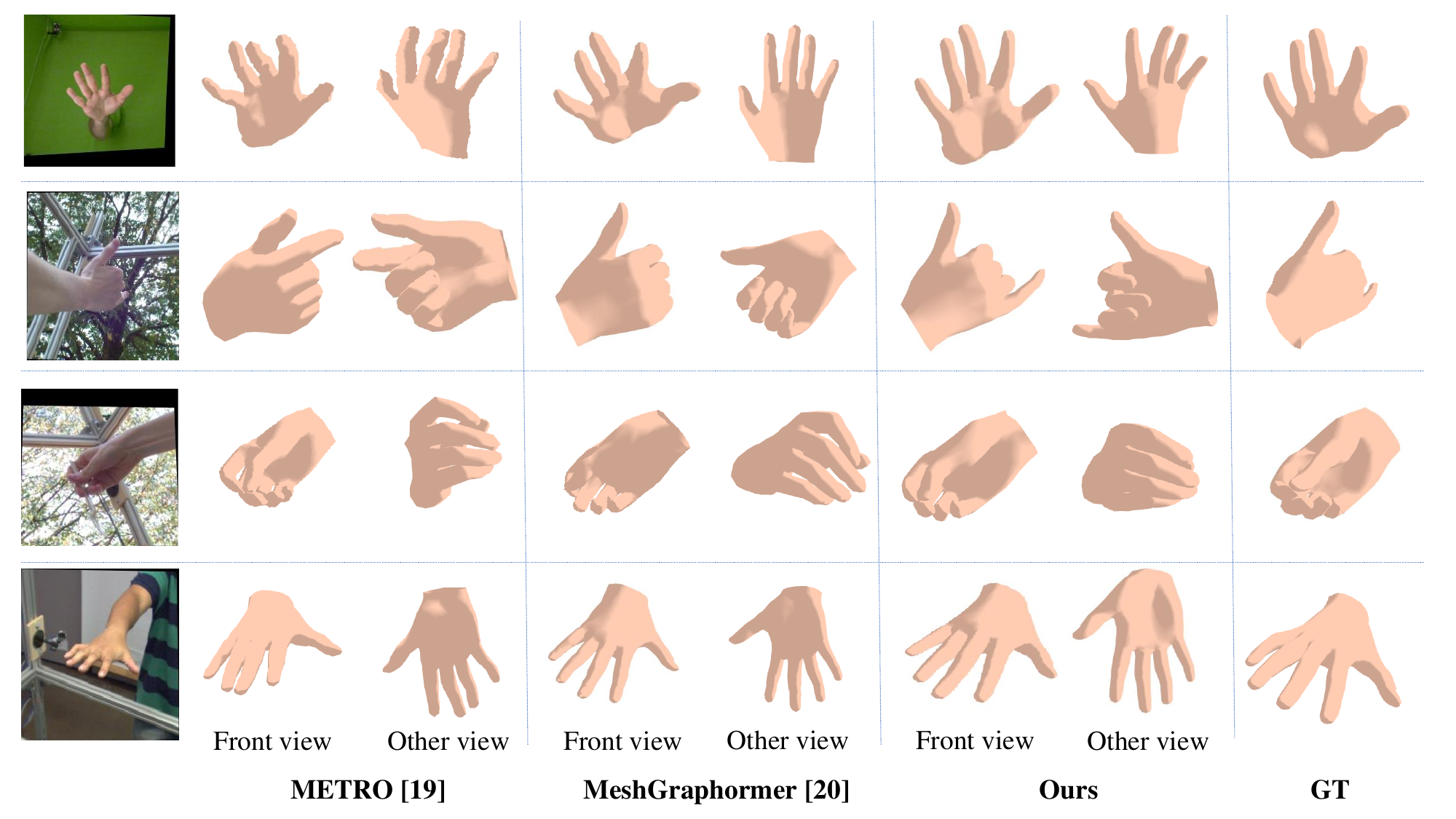}
    \caption{Qualitative comparison of the hand meshes produced by ReJSHand and other state-of-the-art methods.
    Results show that our approach achieves more accurate hand mesh reconstruction outcomes that are closer to the ground truth.}
    \label{fig:comp}
\end{figure*}

\subsection{Datasets}

ReJSHand is tested and evaluated on FreiHAND~\cite{zimmermann2019freihand}. This dataset, developed by the computer vision lab at the University of Freiburg, constitutes a large-scale hand pose and shape estimation dataset designed for training and evaluating deep neural networks using monochrome images. The dataset comprises 130,240 unique training samples and 3,960 unique evaluation samples.

\subsection{Training configuration}

The proposed framework was implemented in Python using the PyTorch deep learning library~\cite{NEURIPS2019_9015}. Its backbone model was FastViT-MA36~\cite{vasu2023fastvit}, pre-trained on ImageNet. During supervised learning, Adam~\cite{kingma2014adam} was used as the optimizer, with a total of 200 epochs for training. The learning rate was set to $5 \times 10^{-4}$ for the first 100 epochs and $5 \times 10^{-5}$ for the next 100 epochs. Finally, ReJSHand was trained on an NVIDIA GeForce RTX 4090 Ti with 24 GB of RAM, and training it took approximately 4 hours.

\subsection{Evaluation metrics}

We evaluate the real-time performance and accuracy of constructing hand meshes using ReJSHand through several metrics: Procrustes-aligned Mean Per Joint Position Error (PA-MPJPE, see Eqn.~\ref{eq4}), Procrustes-aligned Mean Per Vertex Position Error (PA-MPVPE, see Eqn.~\ref{eq5}), Mean Per Joint Position Error (MPJPE), Mean Per Vertex Position Error (MPVPE), F-Score, and frames per second (FPS).

\begin{align}
    PA-MPJPE = \sqrt{\frac{1}{N} \sum_{i=1}^{N} \min_{rigid} ||P_{pred}^{(i)} - P_{gt}^{(i)} \circ rigid||^2_2} \label{eq4}, \\
    PA-MPVPE = \sqrt{\frac{1}{M} \sum_{j=1}^{M} \min_{rigid} ||V_{pred}^{(j)} - V_{gt}^{(j)} \circ rigid||^2_2} \label{eq5},
\end{align}
where $N$ represents the number of joints, and $M$ is the number of vertices. $P_{\text{pred}}^{(i)}$ denotes the predicted position of the $i$-th joint, and $P_{\text{gt}}^{(i)}$ is the corresponding true position of the $i$-th joint. The symbol $\circ$ indicates the application of a rigid transformation, such as rotation, translation, and scaling. $\min_{\text{rigid}}$ refers to finding the minimum value under all rigid transformations, typically achieved through Procrustes analysis.

PA-MPJPE is a commonly used metric for assessing the performance of 3D human pose estimation algorithms. It involves an optimal rigid transformation, known as Procrustes alignment, applied to the predicted poses before computing the MPJPE. This transformation includes rotation, translation, and scaling, which eliminates the effects of global rotation and translation. As a result, the evaluation focuses on the accuracy of the pose structure itself.

\subsection{Comparative results}

\begin{table*}[t!]
    \centering
    \caption{
    The experimental outcomes on the FreiHand dataset, as well as the details of various models, are presented. The proposed pipeline demonstrates strong computational performance, achieving high scores in both frames per second and accuracy. The first and second best results are marked in \textbf{\textcolor{red}{Red}} and \textbf{\textcolor{blue}{Blue}}.
    }
    \label{tab1}
    \begin{tabular}{ccc|ccccc}
    \hline
        Method & Backbone & Backbone Pre-train dataset & PA-MPJPE $\downarrow$ & PA-MPVPE $\downarrow$  & F@05 $\uparrow$ & F@15 $\uparrow$ & FPS \\ 
        \hline
        I2L-MeshNet \cite{chen2021i2uv} & ResNet50 & ImageNet & 7.4 & 7.6 & 0.681 & 0.973 & 36  \\ 
        CMR \cite{chen2021camera}  & ResNet50 & ImageNet & 6.9 & 7.0 & 0.715 & 0.977 & -  \\ 
        I2UV-HandNet \cite{chen2021i2uv}  & ResNet50 & ImageNet & 7.2 & 7.4 & 0.682 & 0.973 & -  \\ 
        Tang \textit{et al.} \cite{tang2021towards} & ResNet50 & ImageNet & 6.7 & 6.7 & 0.724 & 0.981 & 47  \\ 
        MobRecon \cite{chen2022mobrecon}  & DenseStack & ImageNet & 6.9 & 7.2 & 0.694 & 0.979 & \textbf{\textcolor{blue}{80}}  \\
        METRO \cite{lin2021end} & HRNet & COCO & 6.7 & 6.8 & 0.717 & 0.981 & 27  \\ 
        MeshGraphomer \cite{lin2021mesh}  & HRNet & ImageNet & \textbf{\textcolor{blue}{6.3}} & \textbf{\textcolor{blue}{6.5}} & 0.738 & \textbf{\textcolor{blue}{0.983}} & 10 \\ 
        FastMETRO \cite{cho2022cross}  & HRNet & ImageNet & 6.5 & 7.1 & 0.687 & \textbf{\textcolor{blue}{0.983}} & 28  \\ 
        Deformer \cite{yoshiyasu2023deformable}  & HRNet & COCO & \textbf{\textcolor{red}{6.2}} & \textbf{\textcolor{red}{6.4}} & 0.743 & \textbf{\textcolor{red}{0.984}} & 21  \\ 
        %PointHMR \cite{kim2023sampling}   & HRNet & ImageNet & 6.1 & 6.6 & 0.720 & 0.984 & -  \\ 
        SimpleHand \cite{zhou2024simple}  & FastViT-MA36 & ImageNet & 6.4 & 6.7 & \textbf{\textcolor{blue}{0.744}} & \textbf{\textcolor{red}{0.984}} & 67  \\ 
        FastViT \cite{vasu2023fastvit}   & FastViT-MA36 & ImageNet & 6.6 & 6.7 & 0.722 & 0.981 & \textbf{\textcolor{red}{84}}  \\ 
        \hline
        \textbf{Ours}  & \textbf{FastViT-MA36} & \textbf{ImageNet} & \textbf{\textcolor{blue}{6.3}} & \textbf{\textcolor{red}{6.4}} & \textbf{\textcolor{red}{0.756}} & \textbf{\textcolor{red}{0.984}} & \textbf{72}  \\ 
    \hline
    \end{tabular}
    \begin{tablenotes} 
	   \item $\uparrow$/$\downarrow$ indicates that a higher/lower metric value corresponds to better performance, respectively. 
    Our results are highlighted in bold. 
    \end{tablenotes} 
    \centering
\end{table*}

To demonstrate the effectiveness of ReJSHand, we conducted a comparison with other hand pose estimation methods, including I2L-MeshNet~\cite{chen2021i2uv}, CMR~\cite{chen2021camera}, I2UV-HandNet~\cite{chen2021i2uv}, MobRecon~\cite{chen2022mobrecon}, %PointHMR~\cite{kim2023sampling}, 
FastViT~\cite{vasu2023fastvit}, SimpleHand~\cite{zhou2024simple}, and transformer-based approaches such as METRO~\cite{lin2021end}, MeshGraphomer~\cite{lin2021mesh}, FastMETRO~\cite{cho2022cross}, Deformer~\cite{yoshiyasu2023deformable}, as well as the method proposed by Tang \textit{et al.}~\cite{tang2021towards}. The results are presented in Table~\ref{tab1}.
Additionally, we compared the computational efficiency of ReJSHand against several widely-known techniques, as shown in Fig.~\ref{fig:FPS}. For frames-per-second comparisons, we used an NVIDIA 2080Ti GPU to test the pre-trained models, ensuring consistency with prior works.
Moreover, we conducted a qualitative evaluation by comparing the hand meshes generated by ReJSHand with those produced by other state-of-the-art approaches. The qualitative results depicted in Fig.~\ref{fig:comp} reveal that, compared to previous methods, our approach yields more accurate hand mesh reconstruction outcomes, which are closer to the ground truth.

\begin{table}[t!]
    \renewcommand{\baselinestretch}{1.5}
    \centering\setlength{\tabcolsep}{3pt}
    \caption{Parameter Amounts for different methods.}
    \label{tab:para}
    \footnotesize
    \resizebox{\linewidth}{!}{
        \begin{tabular}{c|c|c|c|c}
            \toprule
            \bf{Methods}                 & \bf{METRO \cite{lin2021end}} & \bf{MeshGraphomer \cite{lin2021mesh}} & \bf{FastMETRO \cite{cho2022cross}} & \bf{Ours} \\ \midrule
            \bf{Parameters}             & 102M & 98M & 25M & \textbf{\textcolor{red}{1.91M}}  \\ 
            \bottomrule
    \end{tabular}
    }
\end{table}

Based on the data reported in Table~\ref{tab1} and the visualization results presented in Fig.~\ref{fig:FPS}, The proposed algorithm achieves the optimal balance between accuracy and computational efficiency. Specifically, the PA-MPJPE is 6.3, the PA-MPVPE is 6.4, F@05 is 0.756, and F@15 is 0.984. Although Deformer~\cite{yoshiyasu2023deformable} outperforms ReJSHand in PA-MPJPE, our framework achieves a faster testing speed of 72 frames per second, which is 3.4 times faster than Deformer. Similarly, even though FastViT~\cite{vasu2023fastvit} and MobRecon~\cite{chen2022mobrecon} achieve higher scores in the frames-per-second metric, their PA-MPJPE and PA-MPVPE values are lower than our method.
Moreover, despite being built on transformer-based architectures, METRO~\cite{lin2021end}, MeshGraphomer~\cite{lin2021mesh}, and FastMETRO~\cite{cho2022cross} have extremely low frames-per-second values (\textit{i.e.}, all below 30), classifying them as non-real-time methods. Additionally, their parameter counts are 102M, 98M, and 25M, respectively, whereas ours is only 1.91M (see Table~\ref{tab:para}). This makes our framework lighter than even SimpleHand~\cite{zhou2024simple}, which has 1.93M parameters.

\subsection{Limitation}

To identify the limitations of our pipeline, we attempted to reproduce SimpleHand~\cite{zhou2024simple} but were unable to achieve the performance reported in the paper. However, to ensure fairness, we tested ReJSHand under the same conditions. To reduce the model size, we experimented with other backbone networks, such as RepVit~\cite{wang2024repvit}. When using RepVit, ReJSHand's parameters decreased to 1.03M, but the results were inferior.
We concluded that this was due to the backbone's insufficient feature extraction capability, which led to inadequate representation of joint and skeleton features. Our experiments demonstrated that the end-to-end generation approach and a lighter model architecture generally yield better performance.
%In the future, we plan to explore more lightweight backbones and feature extraction modules that can effectively represent joint and skeleton features.

\section{Conclusions}
\label{sec:conclusions}
This paper introduces ReJSHand, a lightweight network designed for real-time hand pose estimation and mesh reconstruction. ReJSHand features an innovative architecture that integrates a 2D keypoints generator, an expansion block, a feature interaction block, and a 3D keypoints generator. This design enables the network to achieve exceptional computational efficiency while maintaining high accuracy. Furthermore, the incorporation of a multi-head self-attention mechanism and a coordinate attention layer significantly enhances feature representation, resulting in highly precise hand meshes.
Extensive evaluations on the FreiHand dataset demonstrate ReJSHand's superiority. It achieves a processing speed of 72 frames per second while maintaining low PA-MPJPE and PA-MPVPE values. Moreover, the model's remarkable F-scores and real-time processing capabilities highlight its potential for dynamic interaction scenarios in robotics and human-computer interaction.

% \newpage
%\addtolength{\textheight}{-12cm} % This command serves to balance the column lengths

%%%%%%%%%%%%%%%%%%%%%%%%%%%%%%%%%%%%%%%%%%%%%%%%%%%%%%%%%%%%%

%\section*{APPENDIX}

\section*{Acknowledgement}

The authors gratefully acknowledge the support of the National Key Research and Development Program of China (Grant No. 2023YFC3603601).

%%%%%%%%%%%%%%%%%%%%%%%%%%%%%%%%%%%%%%%%%%%%%%%%%%%%%%%%%%%%%

\bibliographystyle{IEEEtran}
\bibliography{ref}

\end{document}